\pdfoutput=1

\documentclass[11pt]{article}

\usepackage{ACL2023}

\usepackage{times}
\usepackage{latexsym}

\usepackage[T1]{fontenc}

\usepackage[utf8]{inputenc}

\usepackage{microtype}

\usepackage{inconsolata}

\usepackage{graphicx}

\title{\LARGE Is Personality Prediction Possible
Based on Reddit Comments?\\[0.5cm]
\Large University Trier\\M.Sc. Natural Language Processing\\[0.5cm]}

\author{Robert Deimann \\
 robert.deimann@gmail.com\\\And
Till Preidt \\
  \And
 Shaptarshi Roy \\
  \And
  Jan Stanicki \\
 s2jastan@uni-trier.de\\}

\begin{document}
\maketitle

\begin{abstract}
In this assignment, we examine whether there is a correlation between the personality type of a person and the texts they wrote. In order to do this, we  aggregated datasets of Reddit comments labeled with the Myers-Briggs Type Indicator (MBTI) of the author and built different supervised classifiers based on BERT to try to predict the personality of an author given a text. Despite experiencing issues with the unfiltered character of the dataset, we can observe potential in the classification.
\end{abstract}

\section{Team roles and responsibility}
For this project, we split the work into different parts.
There is the task of knowledge collection, finding relevant and related papers. This task involved reading the papers and summarizing an overview in the assignment as well as providing the other team members with the knowledge, so informed decisions can be made. This part of work, as well as the data sampling, was the responsibility of Shaptarshi Roy (1643514).\\
The data collection and processing was done by Jan Stanicki (1644014). This contained all steps described in the section Data collection and also writing said section.\\
Robert Deimann (1615725) dealed with the background knowledge of the Myers-Briggs Type Indicator, programmed the classifiers, evaluated them and also contributed to the analysis and conclusion.\\
Finally Till Preidt (1416911) was responsible for the basic analysis, including language distributions within the classes and a bag of words representation for comparing them. Additionally he formulated the introductory and future work sections.

\section{Introduction}
Natural language processing is a vast and varied field and captures many different tasks relating to different aspects of language. Especially with regards to lexicology, semantics and pragmatics, it is not surprising that also the prediction of an author's personality - or at least some characteristics - is possible only based on the respective written statements of this person. Very interesting in this context are social networks, since their contents - mostly comments - can be used to generate psychological profiles of the users and classify them, even though experts of psychology have doubt regarding the reliability of such procedures. Of course, this depends on the number of statements from a respective user: the more data is available, the more meaningful becomes the generated profile. The inspiration for our project came from previous work on this topic, as described in the next section. \\
Our project specializes on the so called Myers-Briggs type indicator (MBTI). This can be seen as a psychological test for characterising statements - such as comments - into four binary classes, so there are 16 classes in total. For this approach, we collected user comments from the social media platform Reddit. In the context of non-automatic / manual analysis by experts, there could be a focus on the used vocabulary (lexicology) and the meaning (semantics) the respective words can have in the covered context (pragmatics). For our project, we use the transformer model ALBERT \cite{DBLP:journals/corr/abs-1909-11942} to classify the Reddit comments.\\ 
An overview on MBTI and the respective approach chosen is explained in the following sections. At this point, it is important to know that our data set is ambiguous, since there is a distinction between MBTI data and non-MBTI data. MBTI data consists of comments that were posted in the context of a subreddit of the MBTI reddit\footnote{\href{https://www.reddit.com/r/MBTI/}{reddit.com/r/mbti/}}. In accordance to the 16 classes, there are 16 MBTI subreddits named after the respective abbreviations to indicate the respective class. In general, we only use comments from users that have posted a minimum of one comment during the last 5 years in one of these subreddits. Therefore, for each user we have the used subreddit as class / label. So the comments from this users posted in one of the subreddits are the MBTI data and comments from the respective users posted in other contexts (besides this topic) are treated as non-MBTI data. Additionally to this distinction, it was necessary to do a general pre-processing of the data, which included for instance the removal of HTML-artefacts. More details regarding the data collection and treatment will follow in section \ref{sec:data}. For the classification itself, there are several techniques which were used to experiment and compare the respective results to figure out the best classification method as described in Section \ref{sec:method}.

\section{Related Work}

Interdisciplinary tasks like personality prediction and implementing algorithms that can capture nuances that are of a social or psychological nature have been of interest in Data Science and Natural Language Processing research for years. The approaches range from tree-based methods to automatic methods using Language Models. In this section, some of the research related to our work is presented. It is important to note, that all researchers highlight language bias  in their data sets as the most commonly used language in social media and forums is English.\\

Abidin et al. (\citeyear{Abidin2020}) propose a Random Forest classifier for personality prediction to simplify employers’ candidate selection based on the candidates’ applications and social media posts by predicting their MBTI type. Their dataset is derived from Twitter and contains 50 tweets each for 8.657 users and their types. The users were found on \href{www.personalitycafe.com}{personalitycafe.com}, a forum that specializes on the topic of MBTI. To enrich the data, users were asked to fill out a questionnaire in which they specify their personality type and then were asked to chat publicly with other users such that different semantic and syntactic features could be extracted to help with the predictions. A Random Forest classifier is a supervised algorithm that is based on decision trees that are trained on different sets of observations. The final prediction of the Random Forest is based on the average prediction of all trees. This can be beneficial as it can counteract overfitting. This method did significantly outperform other established Machine Learning algorithms such as Logistic Regression and Support Vector Machines.\\

A statistical approach is proposed by Amirhosseini and Kazemian (\citeyear{mti4010009}) who use Extreme Gradient Boosting. This algorithm is an improvement to the basic Gradient Boosting algorithm which is classified as greedy. More regularization has been introduced to the models used for Extreme Gradient Boosting to combat the overfitting issue the original Gradient Boosting algorithm entails. Basically, a sequence of weak models is trained on modified data and their predictions are combined through a weighted majority vote that results in the final prediction. The classification task itself was divided into four binary classifiers specialized in each dimension of the personality types. They use the same dataset as Abidin et al. (\citeyear{Abidin2020}). This method performed best in recognizing differences in the introverted and extroverted dimension of the personality types.\\

Keh and Cheng (\citeyear{DBLP:journals/corr/abs-1907-06333}) propose a fine-tuned BERT model pre-trained on sequence classification. They derived their data directly from the \textit{personalitycafe} forum, which is divided into 16 sections for every personality type. The 5.000 most recent posts from these divisions were scraped for the dataset and the type mentions were masked for the training set. A very interesting part of Keh and Cheng’s research (\citeyear{DBLP:journals/corr/abs-1907-06333}) is their attempt to generate personality type-based language. Here, they used a BERT transformer \cite{devlin-etal-2019-bert} pre-trained on the Masked Language Model task. It was more successful in imitating extroverted personality types as they tend to be more vocal and active in this particular forum which means there was more data to train on.\\

Another interesting approach is multi-view multi-row BERT classifier proposed by Sang et al. (\citeyear{https://doi.org/10.48550/arxiv.2210.10994}). The focus of this work is the personality prediction of movie characters which means the input for the modified BERT classifier will be much longer than intended originally by Devlin et al. (\citeyear{devlin-etal-2019-bert}). The inputs are also divided into non-/verbal inputs which have to be handled differently to predict the personality type of the character. These non-/verbal inputs are sourced from scripts derived from the \href{https://imsdb.com/}{Internet Movie Script Database}. These scripts are parsed to collect dialogue and scene descriptions. The \href{https://www.personality-database.com/}{Personality Database} is used to match the character names, which are found by using a Soft Name Matching algorithm, to their personality types. In total, this dataset contains 3.543 characters from 507 movies. Not only do Sang et al. (\citeyear{https://doi.org/10.48550/arxiv.2210.10994}) introduce a BERT classifier that can handle longer inputs, they also propose a statistical parser that can automatically parse different parts of a script to its fundamental parts. Both contributions can be useful and can be expanded to other topics of research.\\

An inspiration for this paper was the work by Gjurkovic and Šnajder (\citeyear{gjurkovic-snajder-2018-reddit}) which focuses on data acquisition to improve personality prediction models. Here, personality prediction is defined as a supervised machine learning task. The problems they claim to resolve with their data set include non-anonymity, limited space for expression (e.g. through character limitations on Twitter) and a bias towards personality-related topics as sources of data were personality forums as seen in the previously mentioned papers. These problems are the reasons why Gjurkovic and Šnajder (\citeyear{gjurkovic-snajder-2018-reddit}) introduce the MBTI data set and a subset, MBTI9k, a benchmark data set labeled with MBTI types and other linguistic and user-related features. The data set is divided by posts and comments: There are 22.934.193 comments (583.385.564 words) from 36.676 subreddits by 13.631 users and 354.996 posts (921.269 words) from 20.149 subreddits by 9.872 users. For the subset, only the users were included who posted comments containing more than a 1.000 words. To further remove topic bias, all comments from 122 MBTI-related subreddits were excluded and mentions of personality types and cognitive functions were replaced by placeholders for the training set. \\
It is the largest, openly accessible data set for this task. They are one of the first researchers that used Reddit as a data source for this purpose. Using Reddit as a source remedied many of the aforementioned problems as the posts and comments are anonymous, many users that frequent personality-based communities self-declare their personality types and there is no limit in how long a post or comment can be. Additionally, users tend to comment and post about different topics which mitigates the bias on personality topics and introduces a varied vocabulary. Using MBTI9k majorly improved existing models for personality prediction in every dimension of the Myers-Briggs Indicators showing that diversity in data content and an increase of data size can be beneficial for this task. The increased size also allows for the application of simpler methods and the use of simpler features (such as n-grams) to be effective.

\section{Myers Briggs Type Indicator}
The Myers-Briggs Type Indicator (MBTI) is a popular personality assessment tool used widely in the field of psychology. It was developed by Katherine Briggs and her daughter Isabel Briggs Myers in the 1950s. Katherine Briggs, a writer and mother, was interested in understanding the differences between people's personalities and how this impacted their lives. She was particularly intrigued by the ideas of Carl Jung, a Swiss psychologist who proposed the theory of psychological types. Katherine Briggs was convinced that understanding these types could help people to better understand themselves and others. Together with her daughter, a psychology graduate, she developed the MBTI model and questionnaire, which builds upon and extends Jung's theory of personality types. Today, the MBTI is used in a variety of contexts, including education, career counseling, and personal development \cite{myers1962myers}. \\

Carl Jung was a Swiss psychiatrist and psychoanalyst who is best known for his work on psychological types \cite{jung1923}. He believed that each individual had a unique way of experiencing the world based on their psychological makeup. Jung proposed the idea that people have two main attitudes - introversion and extraversion - which are reflected in their social behavior, and four main functions - thinking, feeling, sensation, and intuition - which determine how they process information. Combining an attitude (extraversion/introversion) with one of the four main functions (thinking/feeling/sensation/intuition) results in a total of 8 different dominant functions / personality types. He argued that the understanding of the dominant function is crucial to self-awareness and personal growth.  \\
 
The MBTI is an extension of Jung's theory of psychological types, as it further defines and operationalizes the concepts of introversion and extroversion, and implements a 'stack' of four functions rather than only focusing on the dominant function of a person. Jung's work focused on understanding the differences between people's personalities, but it did not provide a clear and concise way of categorizing these differences. MBTI, on the other hand, provides a standardized framework for identifying and understanding these differences, using a four-letter code to distinguish between a total of 16 personality types. It extends Jung's idea by adding a secondary, tertiary and inferior function to an individual's personality type, making the typing more distinguished. The two first functions can be seen as 'pilot and co-pilot'. They identify the strengths and define how a well-functioning being is operating most of the time, whereas the two bottom functions describe more of a hidden aspect of the personality, they are underdeveloped and relate more strongly to a person's weaknesses and give hints on what an individual needs to focus on in order to grow. Each of these functions are either directed inwardly or outwardly (introverted or extroverted), in varying order. \\
The four-letter code breaks down as follows:\\
I/E   N/S   T/F   P/J, which stand for: Introverted/Extroverted, Intuition/Sensing, Thinking/Feeling, and Perceiving/Judging.\\
The first letter defines the direction (I/E) of a person's dominant function, and the last letter (P/J) defines which of the two first functions is directed outwardly. Intuition (N) and Sensing (S) are perceiving functions, whereas Thinking (T) and Feeling (F) are judging functions. Thus, an individual's first extroverted function would be \textit{N} or \textit{S} in case they have a \textit{P} at the end and vice-versa the first extroverted function would be \textit{T} or \textit{F} if they have a \textit{J} at the end. The third function is the opposite of the second function, and the fourth (inferior function) is the opposite of an individual's primary function. This all sounds a bit complicated, thus we break it down here for the two opposite types \textit{ESFJ} and \textit{INTP}:\\

\begin{tabular}{||c c||} 
 \hline
 ESFJ  & INTP\\ [0.5ex] 
 \hline\hline
 Extraverted Feeling & Introverted Thinking\\ 
 \hline
 Introverted Sensing & Extroverted Intuition \\
 \hline
 Extraverted Intuition & Introverted Sensing  \\
 \hline
Introverted Thinking & Extroverted Feeling\\
 \hline
\end{tabular}
\\

Going into detail of what each function means exactly and how the position in the functional stack, as well as the interplay between functions affects an individual's personality, would go beyond the scope of this paper. To put it into easy terms, intuitive (N) individuals perceive the world more via their intuition, in a more abstract way, whereas sensing types (S) are more grounded in the 'real world' and explore the world with their senses. Thinking (T) individuals usually base their decisions in rational thought, whereas feeling (F) individuals will usually trust their gut feeling more when it comes to bigger decisions. The distinguishing between the direction (introversion/extroversion) of the individual functions is nuanced and differs for each function. Someone who is an introverted feeler (\textit{INFP} or \textit{ISFP}) will have a rich and complex inner emotional world and may feel things more deeply than thinking types. Someone who is an extroverted feeler (\textit{ESFJ} or \textit{ENFJ}) directs the feeling outwardly, which means they are usually very empathetic individuals. \\

Whilst MBTI is not fully accepted in academia, it is still used widely, has gathered a big following and according to \cite{cohen1981} it is proven that a friend of a person can accurately guess their MBTI type. Due to this as well as personal experience, we feel that there is validity in the concept of MBTI. \\
We were wondering, whether the MBTI of a person can be guessed just based on some text they wrote, and want to see if there is a correlation between MBTI and written text. For this we aggregated labeled data from Reddit comments and built different classifiers on differently assembled datasets, to try to achieve a well-fitting model. We built classifiers based on the full 16 Types MBTI types, the Jungian 8 dominant functions, as well as based on the first two functions (also results in 8 labels). All of these are approaches that we have not seen in any of the related work, as most existing work is simply trying to implement 4 binary classifiers for the respective letters, which we also do. 
Furthermore, we take our existing classifiers for 16 types and split it down into the respective 8-fold classifiers, as well as the 4 binary classifiers, simply by merging the according MBTI-types and then compare this result to the one achieved by the specialized classifiers.

\section{Data Collection} \label{sec:data}
\begin{figure}[h]
    \centering
	\includegraphics[width=\linewidth]{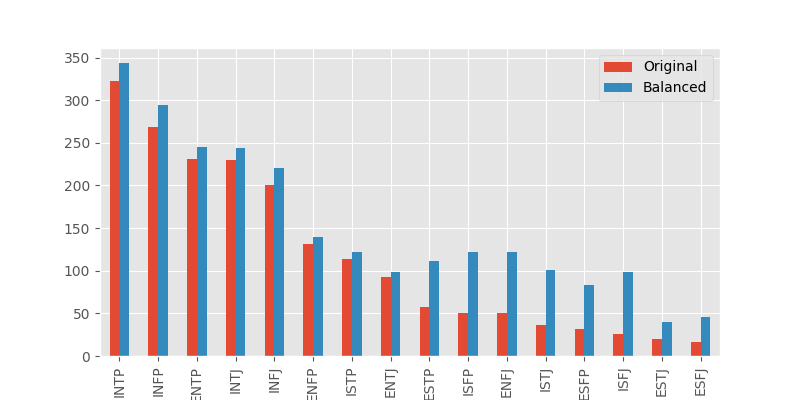}
	\caption{Distribution of users given their label. Red: Number of authors from \texttt{/r/MBTI/}. Blue: Number of authors with additional collection from specific class-subreddits.}
	\label{fig:data:dist_users}
\end{figure}
 
Due to the popularity of the personality test, there are several communities that discuss this topic.
On the social media site \href{www.reddit.com}{www.reddit.com}, there exists a subreddit \texttt{/r/MBTI/}. The users are able to display a flair (here: label) with their respective personality type.
We are able to collect these flairs and utilize them as our classification gold labels. We work with the assumption that the users performed the test and provide the correct label.
To collect the data, reddit does provide an API \textit{PRAW}\footnote{\href{https://praw.readthedocs.io/en/stable/}{PRAW:The Python Reddit API Wrapper}}. Since the API from Reddit limits the number of requests to 1000 at a time and the feature to requests posts in a given time frame is no longer available, we chose to use the more common Pushshift API\footnote{\href{https://api.pushshift.io}{https://api.pushshift.io}}.
The flair needs to be bound to a user themselves meaning that a user can get other flairs in different subreddits. To acquire our class labels, we needed to start collecting users in the MBTI-subreddit, since this is the place we have the connection of a user to a class.\\
We used a two-step process for data collection: retrieving the users and their respective labels and secondly, retrieving the comments of the users across all of reddit. While there is the option to also examine the main post submissions, we choose to focus on the comments because we assume that the behaviour in a conversation contains more data about a personality type. \\
During the retrieval of the users in the subreddit, we collect a total of 1879 users with a corresponding label. However, we observe a significant imbalance of the distribution of users, so we enrich the bottom eight classes with the specific subreddits to find more users with rare personality types. In Figure \ref{fig:data:dist_users} it is shown that the distribution, although still not completely balanced, is more even than before and the sparsely represented classes are now better represented with a total of 2432 users. By filtering by users with a self-proclaimed MBTI-label, we hope to exclude bot- or spam-accounts who cannot display a personality with learnable content.\\
With this approach, we are able to collect the comments of a user across multiple subreddits instead of limiting our data to the \texttt{MBTI} subreddit. The aim is to provide a wider range of topics and contents in the corpus.\\
We collected a total number of around 6.6M comments. In the pre-processing step, we removed around 90k \texttt{'[deleted]'} and \texttt{'[removed]'} comments. In addition to that, we also excluded comments shorter than 50 characters. This is because we assume that short comments are not likely to express the personality of the user or the comment is highly context-dependent as a reaction to a parent-level comment which may not be included in our data. Comments starting with \texttt{'http'} and \texttt{'r/'} were also removed from the corpus because the probability is high that the comments only contain a link or a link to a specific subreddit. With these measures, the dataset ended up with a size of 4.06M comments.\\
Since the base of our user collection are the main \texttt{/r/MBTI/}- as well as the respective type-subreddits, about 630k comments originate from these subreddits and have the Myers-Briggs type indicator test as a subject of discussion. To not give away the information of the user label in comments like \textit{"I am \{insert type\}[\dots]"} to our model, we decided to mask the respective type in the text. This way, the model is not able to learn the correct label from expressions like this but has to find different properties in the text.

\subsection{Limitations of the dataset}
Due to the comments being real-world data, the comments include a lot of spelling mistakes as well as colloquial language, so it results in a more complex dataset. We also have to take into account that the population of users is not distributed comparably to the population of a society. In general, the users are younger and seem to be more prone to a queer spectrum given the active subreddits. An interesting observation from Amirhosseini and Kazemian (\citeyear{mti4010009}) is that the distribution of the general population differs drastically from the distribution in our dataset. In their paper, they show that the most popular types are \textit{ISFJ} and \textit{ESFJ} with 13.8\% and 12.3\% respectively. In contrast, \textit{INTP}, the most frequent type in our dataset (19.4\%)is only represented with 3.3\% in the general population. This indicates that the \textit{INTP} seem to be more eager to contribute in the community and more interested in discussing their personality types than most others. It is also of noteworthy that the rarest class in our dataset (\textit{ESFJ}) is the exact opposite of the most popular class in the dataset. Since different users interact in different communities, there is also a mixture of community or moderation guidelines and different standards of what kind of content is appreciated as a comment.
Irony and sarcasm are also likely to be included in some comments. This reverses the semantic value of the statement which then may lead to false conclusions regarding the expressed personality. The open approach of the data collection also infers the option of low-quality comments in the dataset. Not all comments are well-structured and contain well-written language.

\subsection{Sampling}

\begin{table*}
\centering
\begin{tabular}{rlllll}
\hline
\textbf{Sample} &  & \textbf{\# comments/class} & \textbf{total \# comments} & \multicolumn{2}{l}{\textbf{\# comments/author}} \\
 &  &  &  & \textbf{mean} & \textbf{median} \\
\hline
Total & balanced & 6250 & 100k & 9.65 & 5 \\
 & proportionate & see figure \ref{fig:prop_samples} & 100k & 7.76 & 3 \\
MBTI-only & balanced & 3300 & 52k & 46.53 & 18 \\
 & proportionate & see figure \ref{fig:prop_samples} & 100k & 26.84 & 10 \\
No MBTI & balanced & 6250 & 100k & 6.29 & 3 \\
 & proportionate & see figure \ref{fig:prop_samples} & 100k & 5.38 & 2 \\

\hline
\end{tabular}
\caption{Samples of the collected reddit comments. Total samples from all comments, MBTI-only samples only from MBTI-related comments, No MBTI excludes comments from MBTI-related subreddits.}
\label{tab:samples}
\end{table*}

\begin{figure}[h]
    \centering
    \includegraphics[width=\linewidth]{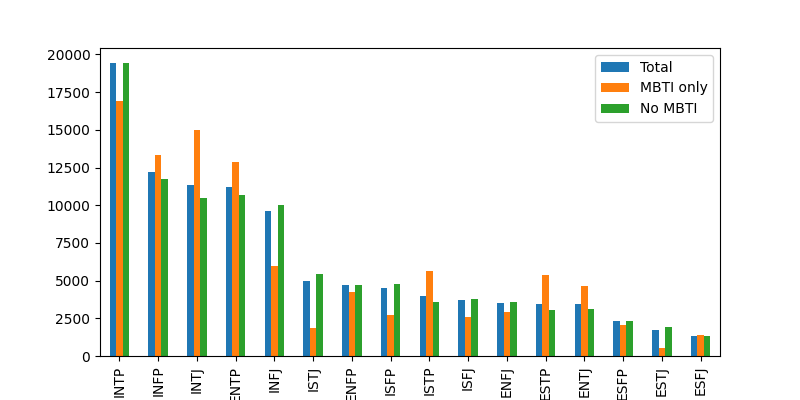}
    \caption{Distribution of classes in the proportionate samples. Blue: Total sample. Green: sample without MBTI-subreddits. Orange: MBTI-only subreddits. }
    \label{fig:prop_samples}
\end{figure}
Considering the amount of data we were able to collect and the available resources, we opted to work with samples. In figure \ref{fig:prop_samples}, the distribution of the comments given their labels is shown. 
We decided to experiment with two approaches of sampling, proportionate and disproportionate sampling. In the disproportionate sampling approach, each class is represented by an equal number of instances. In table \ref{tab:samples}, the number of comments per class are displayed. Due to the limit of the sparsest class in MBTI-only subreddits, all classes were limited to guarantee an equal distribution. \\
The proportionate sampling displays each class in relation to how often it appears. For instance, the class \textit{INTP} as the most common class is represented a lot more than \textit{ESFJ} as the least common class. This is quite an interesting finding, as these two types are exact opposites, which also indicates opposite interests and is reflected in the amount of comments they contributed online. We also see the number of comments per author in table \ref{tab:samples}. The low number of less than ten comments per author shows a high variety of different inputs. That minimizes the possibility of the model picking up user-specific character traits instead of learning personality-specific traits. It also makes sense that the number of comments per author is higher in the MBTI-only samples because the pool of comments is significantly smaller, so an author appears more often.

\section{Method} \label{sec:method}
For the classification, we make use of ALBERT \cite{DBLP:journals/corr/abs-1909-11942} utilizing 'albert-base-v2' for the tokenizer and our base-model. It is known for being efficient with resources while still achieving a solid performance. For the sequence classification we utilize the \textit{AlbertForSequenceClassification} pipeline. We train on different data samples to compare the improvement in predicting the results. For this, we use a learning rate of 2e-5 and AdamW as an optimizer. The number of hidden layers is the default value 12. For the number of training epochs, the value is adapted to find the best generalizing model. \\

As mentioned before, we train different classifiers with the same base structure on different subsets of our big aggregated data set, which contains 4.06M comments.
As the original data set is quite imbalanced, with the most frequent type \textit{INTP} being about 15 times more frequent than the least frequent type in our data set \textit{ESFJ}, we assembled a subset in which each personality type appears at the same frequency. Furthermore, these data sets are further subdivided: a part which only includes comments from the MBTI-specific Reddit subforums, as well as a part which excludes any comments of the MBTI-specific subforums. This last distinction was made in order to see, whether it is easier to predict the personality type of comments in discussion forums that are made explicitly for discussing personalities. \\

In terms of training a model, we had the choice between many different combinations to train our classifier. We had a total of 9 options in regards to the data we train our models on, as we collected 6 subsets of differently sampled data (see table \ref{tab:samples}), as well as in regards to the specific classification problem we are trying to solve, which you find listed in the following:
\begin{itemize}
    \item 16-fold classifier for each individual MBTI
    \item 8-fold classifier for the 8 primary functions (e.g. Introverted Thinking)
    \item 8-fold classifier for types that share the first two functions (e.g. ENFP \& INFP)
    \item 4 binary classifiers along the I/E, N/S, T/F and P/J axes 
\end{itemize}
This results in 6 * 9 = 54 different combinations for us to potentially train on. As we did not have enough computational resources for this, we selected a subset of these combinations and ended up with having trained 18 classifiers in total. To reiterate, as far as model choice goes, we stuck with the same general ALBERT classification model, to get valid comparisons between fine-tuned models of our different datasets and classifiers.\\
Within the model itself, we did experiment with different training set sizes, with most of our classifiers trained on a training set of 64.000 examples with a development set of 16.000 examples. The final test set had all entries that were seen during training filtered out and consisted of 20.000 or 10.000 examples. We also tested training sizes of 10.000 to 20.000 samples. Whilst these worked well on the binary classifiers it was evident that the 8-fold and 16-fold classifiers benefited from a bigger training set. It would have been great to train on even more data, as described in the future work section \ref{sec:future work}, since we gathered a lot more data than we were able to train on. In terms of hyper-parameter tuning, we experimented with different batch-sizes and used a batch size of 10 for the binary classifiers and a batch-size of 32 for the 16-fold MBTI classifier. With a multi-label classification the larger batch size was beneficial, as ideally the model needs to see at least one example of each class during every training step. It was not possible for us to experiment with even larger batch sizes due to GPU RAM limitations. The binary classifiers also learned reasonably well with a smaller batch size of 8 to 10. The results are presented in the next section (\ref{sec:eval}).  \\

All the code used for training can be found in our \href{https://github.com/robookwus/MBTI-Personality-Classification/}{Github repository}
.


\section{Evaluation} \label{sec:eval}

The predictions of our trainers were evaluated on a balanced test set (with equal distribution amongst all types), as well as the imbalanced dataset (original distribution of types in our dataset). \\
Looking at the confusion matrix for the 16-fold or 8-fold classifiers trained on the proportional dataset it is quite evident that these classifiers prefer the types they were mostly trained on:
\begin{figure}[!htb]
    \centering
    \includegraphics[width=\linewidth]{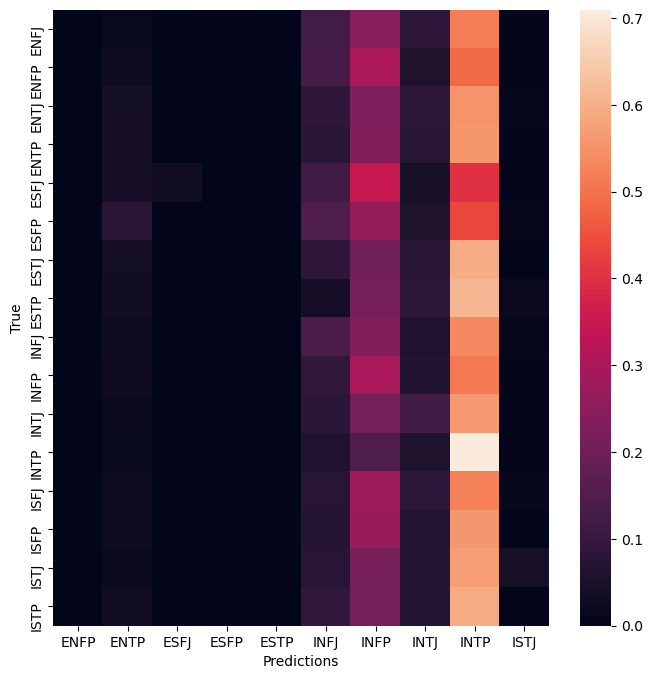}
    \caption{Confusion matrix in early training with imbalanced dataset}
    \label{fig:conf_mat_no_labels}
\end{figure}
After some extended testing, we found the best F1-scores training on the data set sampled to have equal probability for each type. Surprisingly, in the F1-score, the balanced 16-fold classifier outperformed the 16-fold classifier which was trained on the proportional data set (see figures \ref{fig:exp_balanced_heatmap} and \ref{fig:exp_baltrain_baleval_heatmap}. Whilst the F1-score of 16\% , which we achieve with our best 16-fold model does not seem impressive, it has to be noted that we are doing a 16-fold classification, and the value is far above a random distribution, which would have been 1/16 = 6.25\%. It needs to be mentioned that trying to predict the personality type based on some comment snippets without context is also a highly challenging task, leading to way more ambiguous results than e.g. including document classification or sentiment analysis in our research. \\

For the 8-fold classifiers, we merged the opposing two MBTI classes into one class. Whilst we did see an improvement in the prediction accuracy, it was not as significant of a change as we would have liked. We employed two techniques for this: training specifically for the 8-fold classification and simply merging the according types of our 16-fold classification evaluation. Surprisingly, we achieve a better result by merging the results from the 16-fold classification evaluation to an 8-fold classification than we do with the specifically trained models. This may be partially due to not having trained the specific 8-fold models on the balanced data sets, they were only trained on the proportional data sets. The F1-Score for the 8-fold classification is 24\%.    \\

Whilst the last two approaches are uncommon in the existing research and literature known to us, we also explored the most common way of classifying MBTI types: by using four binary classifiers. On each axis (I/E, N/S, T/F and P/J) we achieved F1-scores between 54 and 61\%. Again, merging the 16-fold classifier into the 4 binary classes performed just as well or in some cases even better than the specifically trained models, which is surprising and speaks for our 16-fold classifier. Again, this may be in part due to having used the proportional data set in most cases training the binary classifiers. In retrospect, it would have been better to use the balanced data sets for all the training procedures. In the binary cases, it is apparent that the T/F-axis is the most distinguished yielding the highest F1-score. Our results here are on par with the work of \cite{sang-etal-2022-tvshowguess}. \\

Furthermore, we compared trainers trained on a subset containing only comments from the MBTI-specific subsections of Reddit with comments that where written elsewhere on Reddit. Training on purely the MBTI subset of comments improved the F1-score by 2 percentage points, indicating that personality prediction gets slightly easier in forums dedicated to discussing personalities. Note, that we entirely masked comments containing the acronyms for personality types, so that the model could not learn by simply reading comments containing the personality type.


\section{Analysis}
Before starting to discuss the results of our models, the relation of words (as parts of the comments) and personality types (classes) should be taken into account. Some words can be seen as more representative for a specific class than others. Therefore, we chose the Bag-of-Words-representation (BoW) to get a basis representation of our data before analysing the efficiency of our model.

\subsection{Qualitative Analysis with a Bag of Words approach} \label{sec:qual_analysis}
\begin{figure}[h]
    \centering
    \includegraphics[width=\linewidth]{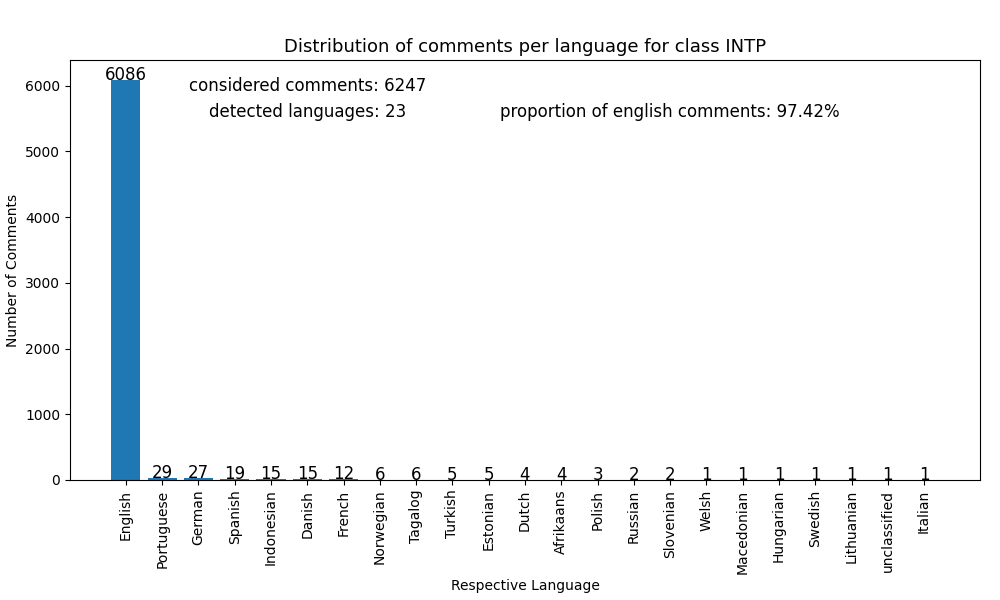}
    \caption{Distribution of languages for the class \textit{INTP} in descending order.}
    \label{fig:prop_language_INTP}
\end{figure}

 The BoW method requires some pre-processing steps, like stemming and the removal of stop words. Unfortunately, these methods are language dependent. Additionally, we had to deal with different languages the comments are written in. So in general that was a ubiquitous problem in our work. That is why it seemed logical to get an overview of all these languages and calculate a respective distribution for the classes. Although, it is interesting to compare all 16 classes this way, it would go beyond the frame of this paper. For that reason, we chose only the following two classes for a comparison: \textit{INTP} (Introvert Intuitive Thinking Perceiving) and \textit{ESFJ} (Extrovert Sensing Feeling Judging), which are opposite types. Furthermore, \textit{INTP} is the most frequent class, while \textit{ESFJ} has the lowest proportion in our data indicating that these types have opposite interests.
For the Analysis, it is important to consider, that the following methods are based on our limited balanced data set. That means, the proportion of only these comments is not significant for the distribution of the types in the real world. Furthermore, it is suggested, that all the specific types of users are equally interested in discussing their personalities on Reddit. Of course, this impression does not represent the real-world proportions of these types' interests, since the willingness to share something depends on characteristic aspects, which logically are not equal between the different types. As already mentioned, the first analysing step was to calculate a distribution of all detected languages for the classes, as shown in figure \ref{fig:prop_language_INTP} for \textit{INTP}.

\begin{figure}[h]
    \centering
    \includegraphics[width=\linewidth]{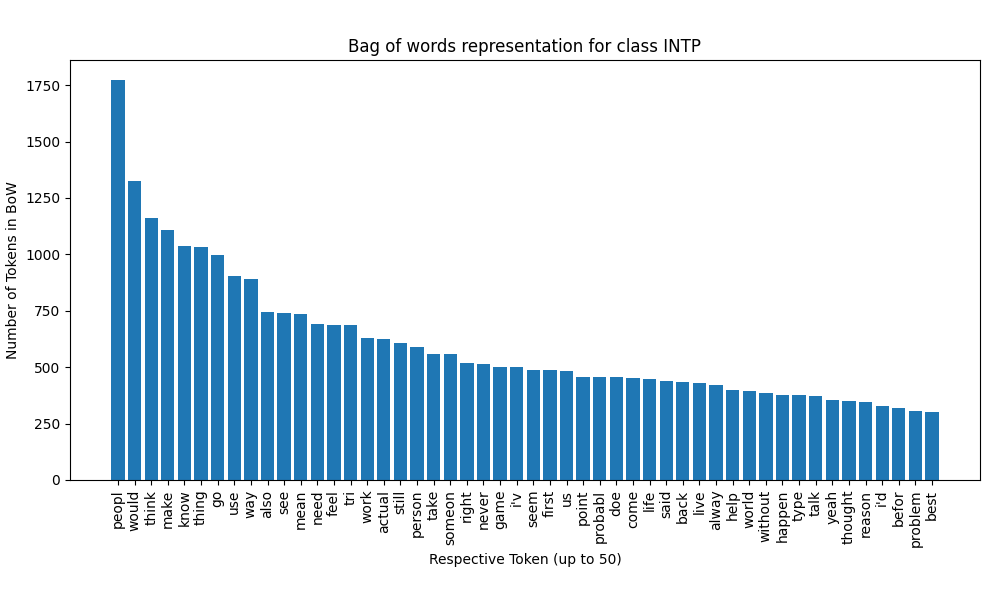}
    \caption{Distribution of languages for the class \textit{INTP} in descending order.}
    \label{fig:BoW_INTP}
\end{figure}

\begin{figure}[h]
    \centering
    \includegraphics[width=\linewidth]{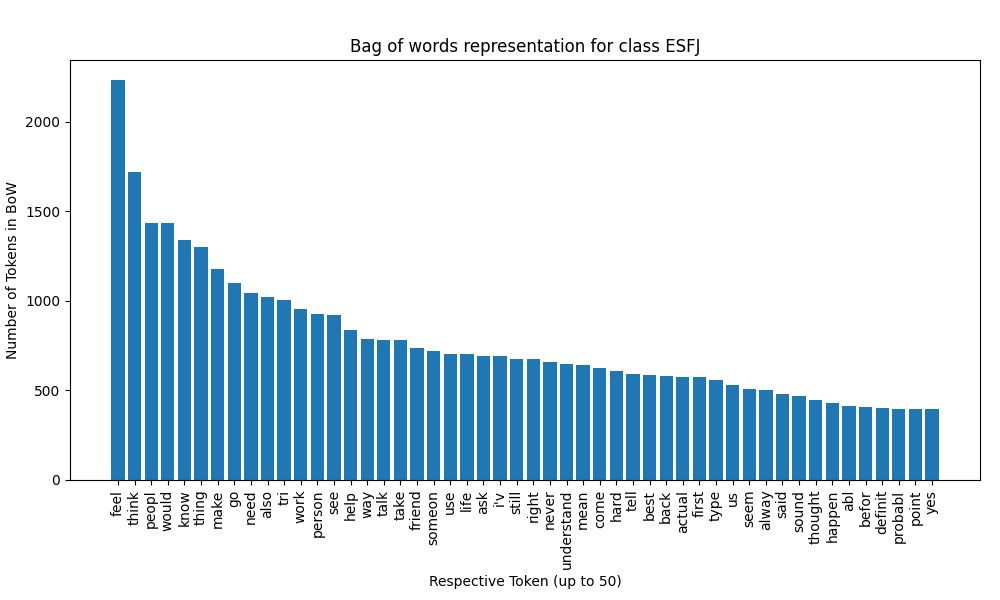}
    \caption{Distribution of languages for the class \textit{ESFJ} in descending order.}
    \label{fig:BoW_ESFJ}
\end{figure}

The calculation of this distributions and the following BoW analysis are both implemented in the script \texttt{lang\_distribution\_and\_bow.py} which is part of the analysis tools folder in the \href{https://github.com/robookwus/MBTI-Personality-Classification/}{Github repository} of our project. The figure shows that in both data sets the amount of the English comments is over-represented. This distribution is similar for all other classes, also for \textit{ESFJ}. Therefore, for BoW it should make no difference to neglect the other languages since the procedure of stemming and stop word extraction can be simplified if there is only one language to deal with.
So in the related script, the focus is always set on the English comments, not depending on which classes are compared. Furthermore, we had to deal with language-independent comment parts like emojis, HTML fragments and complete hyper-reference links. Elements like this were removed via manual functions, while for stemming and stop word extraction specific \textit{nltk} modules were used.
The figures \ref{fig:BoW_INTP} and \ref{fig:BoW_ESFJ} show the Bag-of-Words representation for the respective classes to get a basic impression of which words are relevant for which personality type. 

Regarding the \textit{INTP} class highly represented words like "people", "would", "think", "make" and "know" attract attention since they could be used in polite requests or self-related formulations or in intuitive expressions, on the one hand, and show some kind of introversion on the other hand. In contrast to that, for the \textit{ESFJ} type there are words like "feel", "think", "people", "would" and "know" with high proportion, which shows aspects of extroversion but also sense and feeling. Of course, this data could be more representative since there is only the limited part of our data set taken into account and even the whole data set could capture more comments. Further considerations and possible improvements will be discussed in section \ref{sec:future work}.

\subsection{Classification Analysis}
In the appendix \ref{sec:app_heatmaps}, the heat maps of the results are displayed. We discover that training on a balanced data set (Figure \ref{fig:exp_balanced_heatmap}) improves the performance compared to training the model on the proportionate data set. Here, the data set is evaluated on the proportionate class distributions. In the balanced training, the rarest class \textit{ESFJ} is represented more frequent than in the proportionate sample. This is why the classification works well in comparison (0.29 accuracy of the label). Figure \ref{fig:exp_baltrain_baleval_heatmap} shows the results of the evaluation on the balanced dataset. Although the model seems to lack a confident prediction ability, a strong tendency is apparent. As seen in figure \ref{fig:exp_proportionate_heatmap}, the major classes \textit{INTP}, \textit{INTJ} and \textit{INFP} are predicted most often after training the model on a proportionate data set. The proportionate distribution of the classes is too extreme for the model to recognize the rarer classes. \\
In general, we also have to take into account that predicting the personality based on a single comment is a tough task. It is thoroughly possible that two opposite personalities write the exact same comment or a very similar comment. In section \ref{sec:qual_analysis}, we see that even though there are differences of word choice depending the given class, the majority of frequent words are used by most classes.\\
One quite interesting observation when looking at our confusion matrix heat maps is that each type is misclassified the least as its respective opposite type (e.g. \textit{ENFJ} and \textit{ISTP}. Which indicates that the model also recognizes the opposite types as very different and underlines the validity in the concept of MBTI, where opposite types have the most different personalities. It also shows that, at least to some extent, text can be used to guess the personality of its author.

\section{Future Work}
\label{sec:future work}
A typical aspect of machine learning projects is that there are always many possibilities regarding fine-tuning of the respective models for achieving better results.
Starting with the general idea of changing involved variables like the learning rate, the optimizer or trying several activation functions, the model architecture itself could be adjusted specifically, depending on the used hardware and the desired time expenditure. Of course, the bigger the respective model is, the larger the time loss regarding training and evaluating. On the other hand, using a more complex model has various advantages. 
First of all, it could be useful to capture also the context of a comment and eventually use this context information to distinguish between statements that are meant seriously and those which are meant ironically in a humorous context. Even if the approach would still consist of focusing on comments without context, a more complex model would eventually not be limited to 512 tokens, like the BERT model is in our case. This way, at least the considered dataset could be larger. The time expenditure problem could be minimized by multi-GPU training for more efficiency, which would also allow us to take more data into account at the same time.\\
We were only able to train on a small subset of our overall data, and it would be great to train on all the data to achieve a better model, as gathering this big new data set was a nice addition in MBTI research as well, as most existing work is focusing on the same corpora. 

Another approach could be to focus on optimizing the pre-processing steps. For instance, at this time there is no implementation to deal with emojis, various special characters, different comment length and different languages. Additionally the proportional data sets could be more balanced to achieve a compromise between completely equal sizes of the classes and the proportionality of frequency.
Regarding the classification task, there could be a focus on multiple or all comments of a respective user instead of classifying single comments. This would require a user-based distinction at a previous stage.\\
Another great thing to include in the model would also be context. Personality is hard to predict based on text snippets taken out of context. If a model could make use of the context of a conversation and see how the user reacts to a certain post we believe the accuracy of the predictions could be improved further, as how a person reacts to a certain situation is a big part of an individual's personality as well.
Overall, there several possibilities for experimenting on different aspects of this project.  

\section{Conclusion}
The assignment deals with the question whether it is possible to predict personality types based on comments that were collected on the social media platform Reddit. To achieve this, we collected a number of users from MBTI-related subreddits with self-proclaimed personality type classification. Then, we collected the comments of these users to get a broad representation of personalities. With an ALBERT classifier, we tried to classify the comments of the users. While we encountered several issues in the data collection and classification process, we were able to observe a significant tendency of the model towards the correct prediction. It is questionable if all aspects of personality can even be captured by text snippets only, as there are many more factors that form a persons personality than what they write online. Our work leads us to believe that there are certainly properties that are learnable which can be further evaluated in future work. Our 16-fold MBTI-classification model can be found at \href{https://huggingface.co/JanSt/albert-base-v2_mbti-classification}{huggingface.co/JanSt/albert-base-v2\_mbti-classification}.


\bibliography{anthology,custom}
\bibliographystyle{acl_natbib}
\clearpage
\appendix

\section{Code}
\label{sec:Code}
We exclusively used Python to collect the datasets, implement the classifiers and perform the evaluation and analysis on our data. All the code can be found in our Github repository at \href{https://github.com/robookwus/MBTI-Personality-Classification}{https://github.com/robookwus/MBTI-Personality-Classification}

\section{Results} \label{sec:app_heatmaps}

\begin{figure}[h]
    \centering    \includegraphics[width=\textwidth]{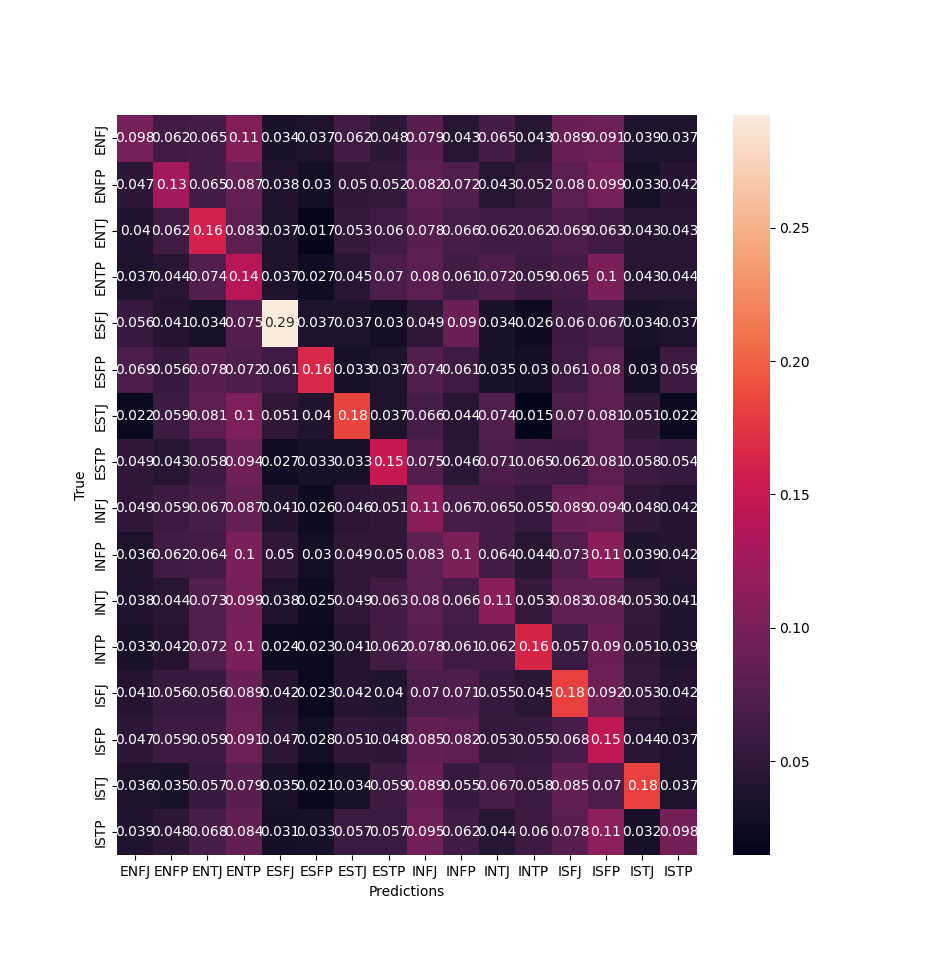}
    \caption{Classification of the personality types divided in 16 classes and trained on a balanced dataset and evaluated on a proportionate dataset sample. }
    \label{fig:exp_balanced_heatmap}
\end{figure}

\begin{figure*}[h]
    \centering
    \includegraphics[width=\linewidth]{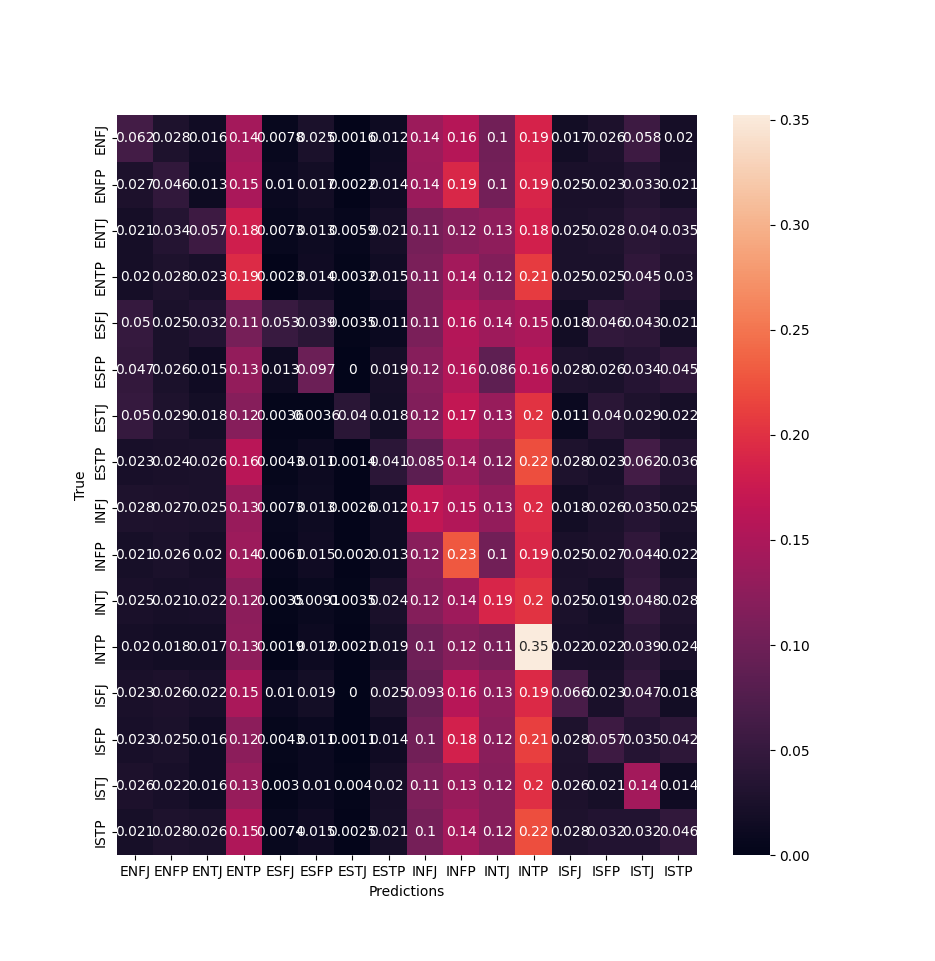}
    \caption{Classification of the personality types divided in 16 classes trained and evaluated on the proportionate dataset sample. }
    \label{fig:exp_proportionate_heatmap}
\end{figure*}

\begin{figure*}[h]
    \centering
    \includegraphics[width=\linewidth]{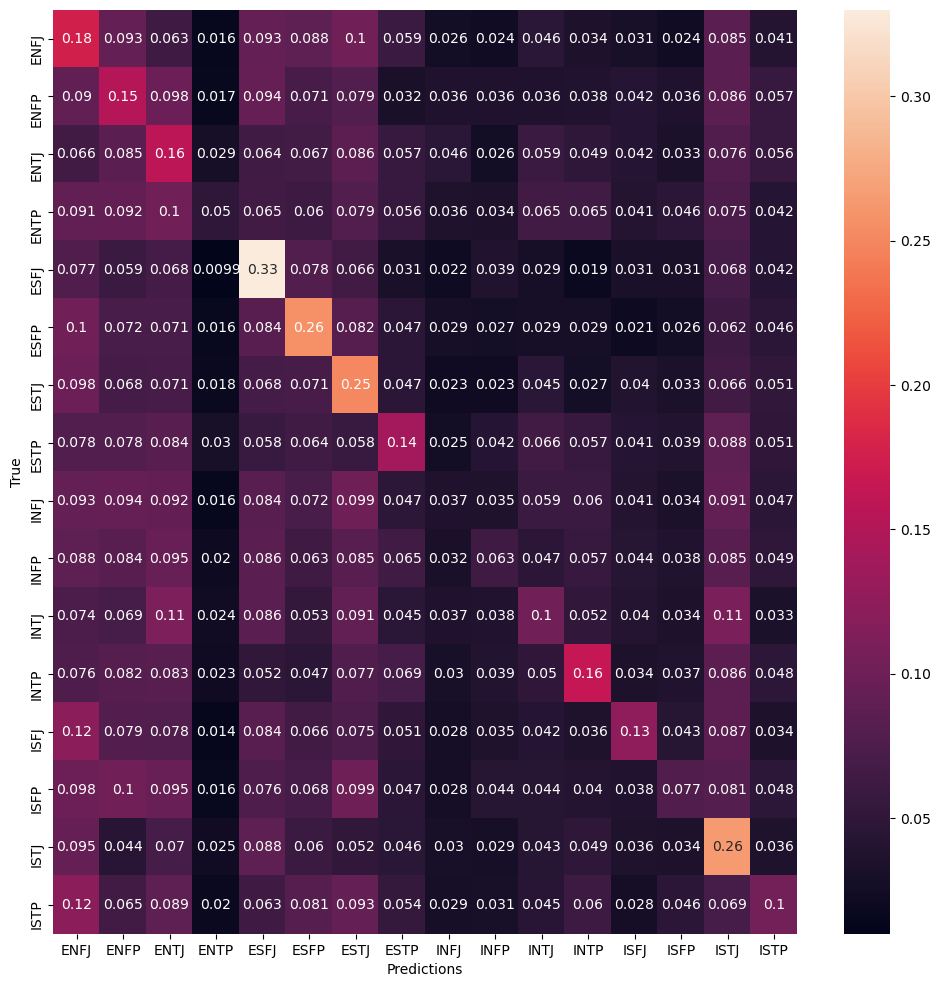}
    \caption{Classification of the personality types divided in 16 classes trained and evaluated on a balanced dataset sample. }
    \label{fig:exp_baltrain_baleval_heatmap}
\end{figure*}

\end{document}